\pdfoutput=1

\documentclass[11pt]{article}

\usepackage[]{acl}

\usepackage{times}
\usepackage{latexsym}

\usepackage[T1]{fontenc}

\usepackage[utf8]{inputenc}

\usepackage{microtype}

\usepackage{graphicx}
\usepackage{algorithm}
\usepackage{algorithmic}
\usepackage{amsmath}
\usepackage{amsthm}
\usepackage{amssymb}
\usepackage{multirow}

\title{\textit{INFINITY}: A Simple Yet Effective Unsupervised Framework for Graph-Text Mutual Conversion}

\author{Yi Xu \quad Luoyi Fu \quad  Zhouhan Lin \quad  Jiexing Qi \quad Xinbing Wang\\
         Shanghai Jiao Tong University \\}

\begin{document}
\maketitle
\begin{abstract}
Graph-to-text (G2T) generation and text-to-graph (T2G) triple extraction are two essential tasks for constructing and applying knowledge graphs. Existing unsupervised approaches turn out to be suitable candidates for jointly learning the two tasks due to their avoidance of using graph-text parallel data. However, they are composed of multiple modules and still require both entity information and relation type in the training process. To this end, we propose \textit{INFINITY}, a simple yet effective unsupervised approach that does not require external annotation tools or additional parallel information. It achieves fully unsupervised graph-text mutual conversion for the first time. Specifically, \textit{INFINITY} treats both G2T and T2G as a bidirectional sequence generation task by fine-tuning only one pretrained seq2seq model. A novel back-translation-based framework is then designed to automatically generate continuous synthetic parallel data. To obtain reasonable graph sequences with structural information from source texts, \textit{INFINITY} employs reward-based training loss by leveraging the advantage of reward augmented maximum likelihood. As a fully unsupervised framework, \textit{INFINITY} is empirically verified to outperform state-of-the-art baselines for G2T and T2G tasks.
\end{abstract}

\section{Introduction}

\begin{figure}[]
    \centering
    \includegraphics[width=1.0\linewidth]{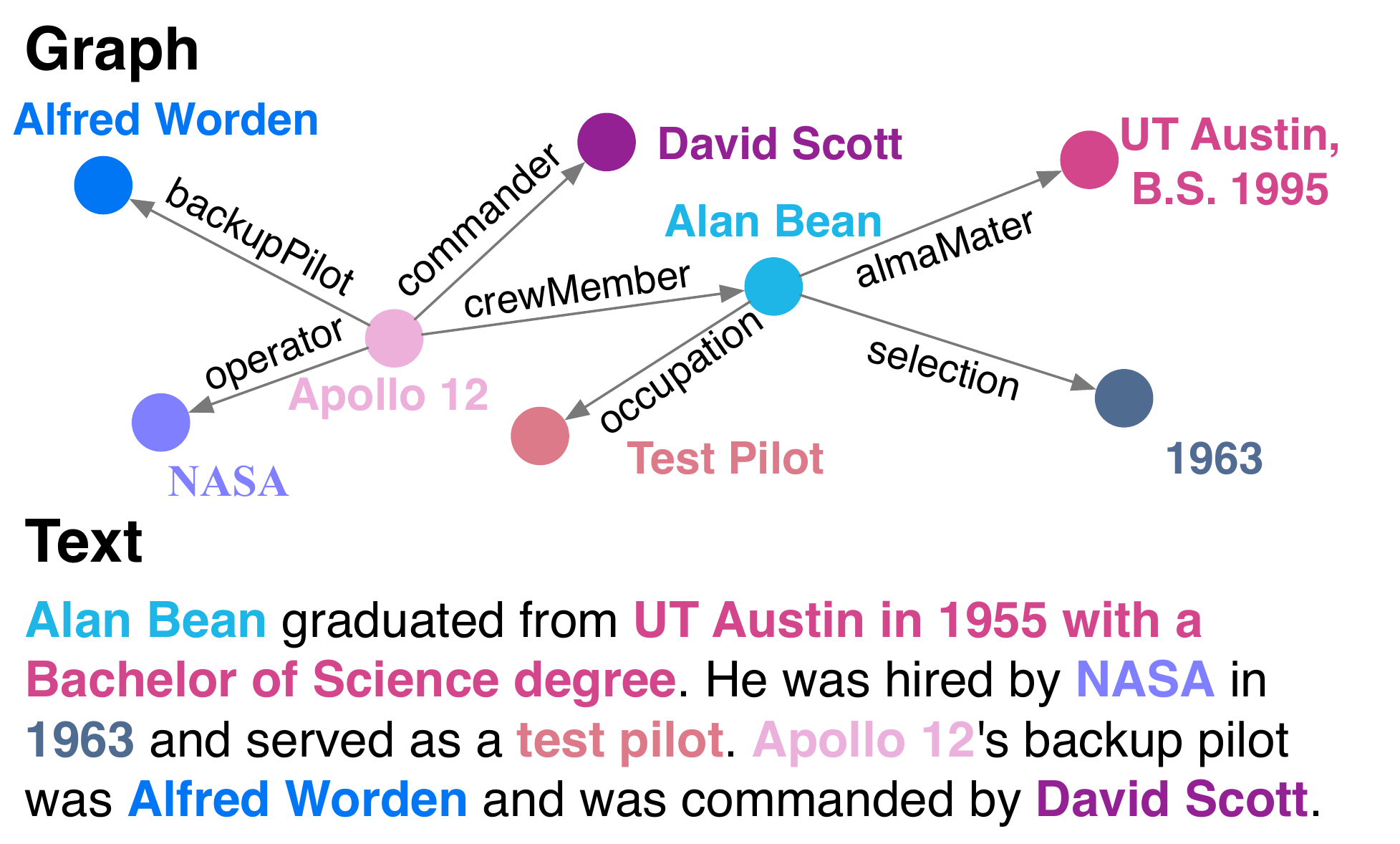}
    \caption{A pair of knowledge subgraph and its corresponding text.}
    \label{fig:graph_text}
\end{figure}

Graph-to-text (G2T) generation and text-to-graph (T2G) triple extraction are two mutually inverse tasks that are crucial to the domain of knowledge graphs (KGs). G2T verbalizes the structural information in KG with descriptive texts, which has attracted much attention to expand the application scope of KG, such as KG-based dialogue and Q\&A system~\citep{ji2021survey}. As a primary task of information extraction, T2G aims to extract triples from text, the typical subtasks of which include named entity recognition (NER) and relation extraction (RE). Figure~\ref{fig:graph_text} illustrates a training pair sample containing part of a knowledge graph and its corresponding text.

\begin{figure}[]
    \centering
    \includegraphics[width=1.0\linewidth]{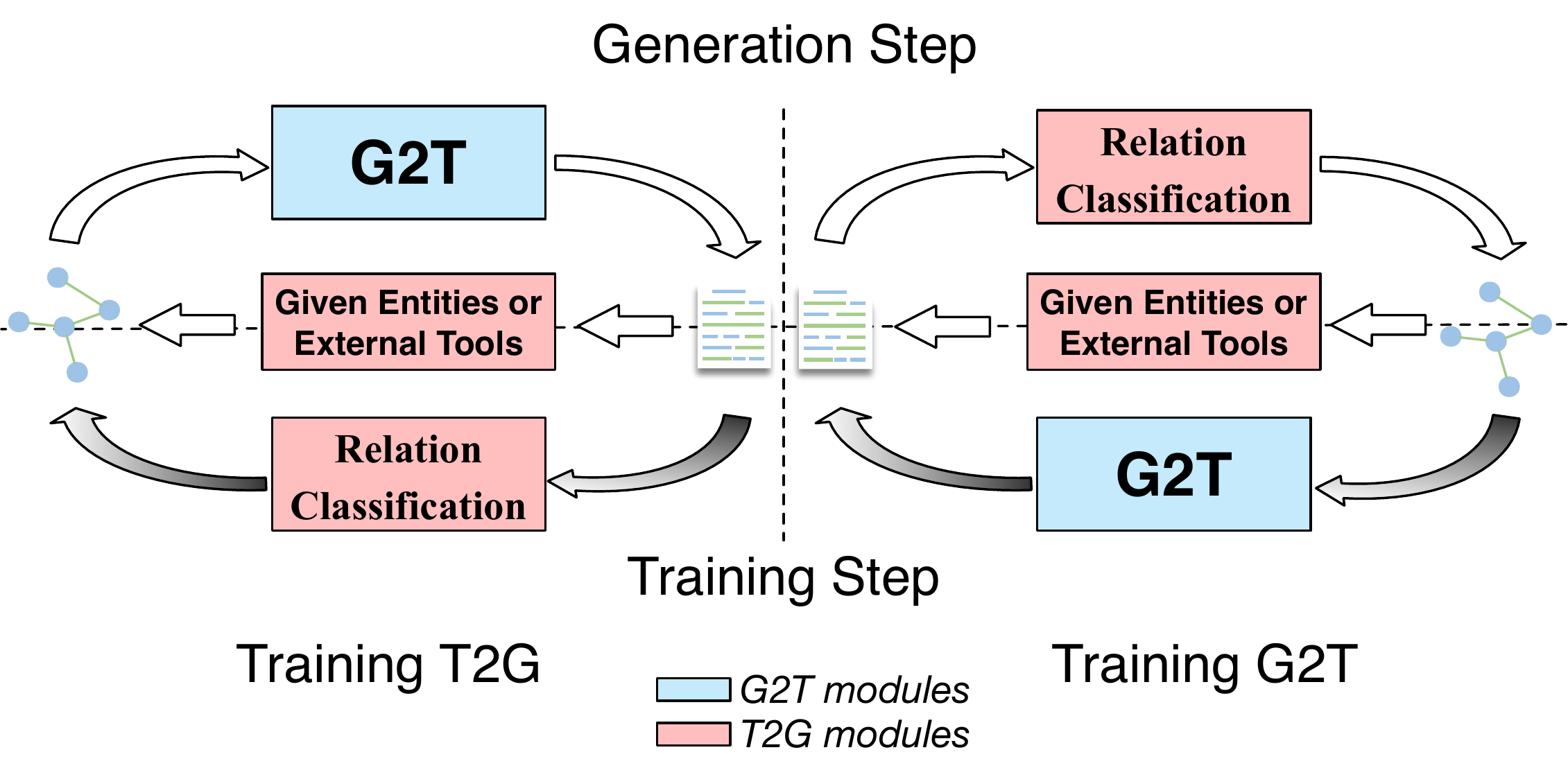}
    \caption{Framework of existing unsupervised models. The left part is the cycle of training T2G, and the right part is the cycle of training G2T.}
    \label{fig:simple_g2t2g}
\end{figure}

G2T and T2G have been intensively studied respectively, mainly treated as two kinds of independent problems in a supervised way. Due to the success of pretrained language models (PLMs)~\citep{raffel2019exploring,lewis2019bart}, mainstream supervised methods have achieved considerable performance with fine-tuning or prompt learning paradigm~\citep{ribeiro2020investigating,clive2021control,ye2021contrastive,ke2021jointgt}. However, these supervised methods require annotated data. Inspired by unsupervised machine translation approaches~\citep{lample2018phrase}, recent work attempts to explore low-resource alternatives that avoid the requirement of graph-text pairs with unsupervised joint learning~\citep{schmitt2020unsupervised,guo2020cyclegt}. As illustrated in Figure~\ref{fig:simple_g2t2g}, unsupervised methods consist of G2T modules and T2G modules with different parameters, which are trained jointly in an iterative manner through the two steps of back-translation: the generation step and training step. The outputs of the generation step for the current modules serve as the supervised training signal for the other modules in the next iteration. Such an interactive and coupling process imperceptibly produces a lot of synthetic parallel data that is helpful to low-resource training.

In this paper, we are thus motivated to focus on unsupervised learning of both G2T and T2G tasks in a joint framework. As shown in Figure~\ref{fig:simple_g2t2g}, existing state-of-the-art models share two major issues in order to be jointly trained. First, current unsupervised models usually simplify the T2G task into relation classification with given entities~\citep{jin2020genwiki}. As a result, the text corpus has to seek external information extraction tools for the acquisition of entity annotations. Second, existing research branches on either G2T or T2G separately implement the two tasks using different neural modules, i.e., G2T modules and T2G modules, which contain numerous parameters that make it challenging to train and share information with each other~\citep{schmitt2020unsupervised,guo2020cyclegt}.

To tackle the above issues, we design a novel back-translation-based framework called \textit{INFINITY} that integrates G2T and T2G tasks under the unsupervised setting. Note that we name our framework as \textit{INFINITY} since the overall architecture of the interaction between G2T and T2G resembles the shape of $\infty$ (Figure~\ref{fig:infinity}). We first investigate the power of seq2seq-based PLMs for G2T and T2G generation and propose to regard graph-text mutual conversion as two sequence generation tasks, where we manage to ensure the simultaneous generation of continuous synthetic pairs of graph-text sequences in a PLM-based module with our newly designed back-translation technique. In this way, \textit{INFINITY} requires no additional neural networks beyond the PLM. To prevent the possible performance deterioration caused by graph linearization, we adopt the reward augmented maximum likelihood~\cite{norouzi2016reward} for training losses to retain the order and structural information in the original dataset during the training process. In contrast to prior unsupervised work~\citep{schmitt2020unsupervised,guo2020cyclegt}, \textit{INFINITY} is entirely bootstrapped without the assistance from manual or automatic annotation tools.

We perform extensive experiments on two benchmarks: WebNLG~\citep{gardent2017webnlg} and GenWiki~\citep{jin2020genwiki}, both of which belong to the very few benchmarks that can evaluate G2T and T2G jointly. The results show the superiority of \textit{INFINITY} over existing methods. Thanks to its simplicity and efficiency, \textit{INFINITY} can be quickly deployed on various scenarios for application. This work presents the following contributions:
\begin{itemize}
    \item We are the first to take G2T and T2G as two sequence generation tasks and propose \textit{INFINITY}, a novel unsupervised framework for graph-text mutual conversion.
    \item \textit{INFINITY} uses only one pretrained seq2seq model to generate synthetic parallel data iteratively and employs the reward augmented maximum likelihood for training loss to obtain structured graph sequences.
    \item \textit{INFINITY} requires no parallel information or external annotation tools compared with other unsupervised models.
    \item We conduct extensive experiments to evaluate \textit{INFINITY} on two benchmarks. The results demonstrate its superiority.
\end{itemize}

\section{Related Work}
\subsection{Supervised Graph-text Models}
As part of the data-to-text task, the key of G2T lies in capturing structural information and generating fluent texts. Some researchers~\citep{marcheggiani2018deep,koncel2019text} design sophisticated architecture based on graph neural networks to encode KGs. In addition, most methods linearize the graph to sequence as input to models. However, graph linearization may lead to the loss of structural information. Researches~\citep{moryossef2019step,zhao2020bridging,guo2020cal} introduce different neural \textit{planner} to determine the order of input triples before linearization. Recently, \citet{ribeiro2020investigating} investigate different PLMs for G2T generation. \citet{clive2021control} propose control prefixes as prompt for PLM, which empowers the model to have finer-grained control during text generation.

Regarding T2G, it aims to extract entities and relations (triples) from given texts, which is usually handled as a classification (tagging) problem to label roles for different tokens~\citep{wei2020novel,yan2021partition}. Apart from these approaches, there emerge some triplet-generation models. CopyRE~\citep{zeng2018extracting} uses the idea of copy mechanism for triple extraction. CPC~\citep{ye2021contrastive} utilizes contrastive learning for direct graph sequence generation, which is similar to the problem definition of \textit{INFINITY}.


\subsection{Unsupervised Graph-text Models}
As shown in Figure~\ref{fig:simple_g2t2g}, unsupervised models combine G2T and T2G into joint learning frameworks. Graph-Text Back Translator (GT-BT)~\citep{schmitt2020unsupervised} is the first approach to unsupervised text generation from KGs and can be used for semantic parsing simultaneously. CycleGT~\citep{guo2020cyclegt} is another unsupervised training method that uses non-parallel graph and text data and iteratively back translates between the two forms. Although GT-BT and CycleGT employ back-translation for unsupervised settings, they simplify the T2G task to relation classification with given entities~\citep{jin2020genwiki}, which requires the text corpus to have entity annotations with external information extraction tools. To some extent, these methods leak the information of parallel corpus in the training process.

\section{Method}\label{sec3}
This section introduces the proposed method \textit{INFINITY}. We first define the tasks and notations. Then we describe the framework and implementation details in the following parts.

\subsection{Formulation and Notations}

Given two non-parallel datasets: a text corpus $\mathcal{T}=\{t_i\}_{i=1}^{N}$, and a graph dataset $\mathcal{G}=\{g_j\}_{j=1}^{M}$, where $N$ and $M$ are the numbers of text sequences and graphs, respectively. Each text sequence in $\mathcal{T}$ can be denoted as $t=(w_1,\cdots,w_L)$ with $L$ tokens, $w_i \in \mathcal{V}$ is the $i$-th token in $t$, and $\mathcal{V}$ is the vocabulary. Each graph in $\mathcal{G}$ consists of a set of triples, denoted as $g=\{(e_h, r, e_t)|e_h,e_t \in \mathcal{E}, r \in \mathcal{R}\}$, where $\mathcal{E}$ and $\mathcal{R}$ represent the entity set and relation type set, respectively. Each entity $e \in \mathcal{E}$ is composed of several tokens formulated as $e=(w_1^e, \cdots, w_{L_e}^e), w_i^e \in \mathcal{V}$. Each relation type $r \in \mathcal{R}$ is also made up of several tokens formulated as $r=(w_1^r, \cdots, w_{L_r}^r), w_i^r \in \mathcal{V}$. Similar to multilingual neural machine translation, we assume $\mathcal{T}$ and $\mathcal{G}$ share the same distribution of latent content $z$ such as linguistic or semantic characteristics:

\begin{align}
    p(g) = \int_{z}p(g|z)p(z)dz,
\end{align}

\begin{align}
    p(t) = \int_{z}p(t|z)p(z)dz,
\end{align}
which is the key of unsupervised learning.
In our unsupervised framework, both G2T and T2G are regarded as sequence generation tasks. G2T aims to generate a natural language text sequence from a knowledge subgraph, while T2G generates a triple sequence that describes the linearized graph where entities and relations exist in the given text. Since the graph itself is a set of triples, for a graph $g \in \mathcal{G}$, we adopt linearization strategy by concatenating all triples with special tokens $[H], [R], [T]$, and $[E]$ to specify the head entity, relation type, tail entity, and end of sequence respectively. The linearized graph is illustrated as follows:

\begin{equation}
    \begin{aligned}
    &[H] \; e_h^1 \; [R] \; r^1 \; [T] \; e_t^1\\
    &[H] \; e_h^2 \; [R] \; r^2 \; [T] \; e_t^2\\
    & \quad \quad \quad \cdots\\
    &[H] \; e_h^{\vert g \vert} \; [R] \; r^{\vert g \vert} \; [T] \; e_t^{\vert g \vert} \; [E],
    \end{aligned}
\end{equation}
where $e_h^i, r^i$, and $e_t^i$ refer to the elements of the $i$-th triple in $g$. We simply linearize the graph using the order of triples in the original dataset. Note that we do not consider other sophisticated methods for linearization since these methods will lead to additional neural components, and we only focus on the proposed framework rather than the neural components inside.

\subsection{Joint Training Framework of G2T \& T2G}
\begin{figure}[t]
    \centering
    \includegraphics[width=1.0\linewidth]{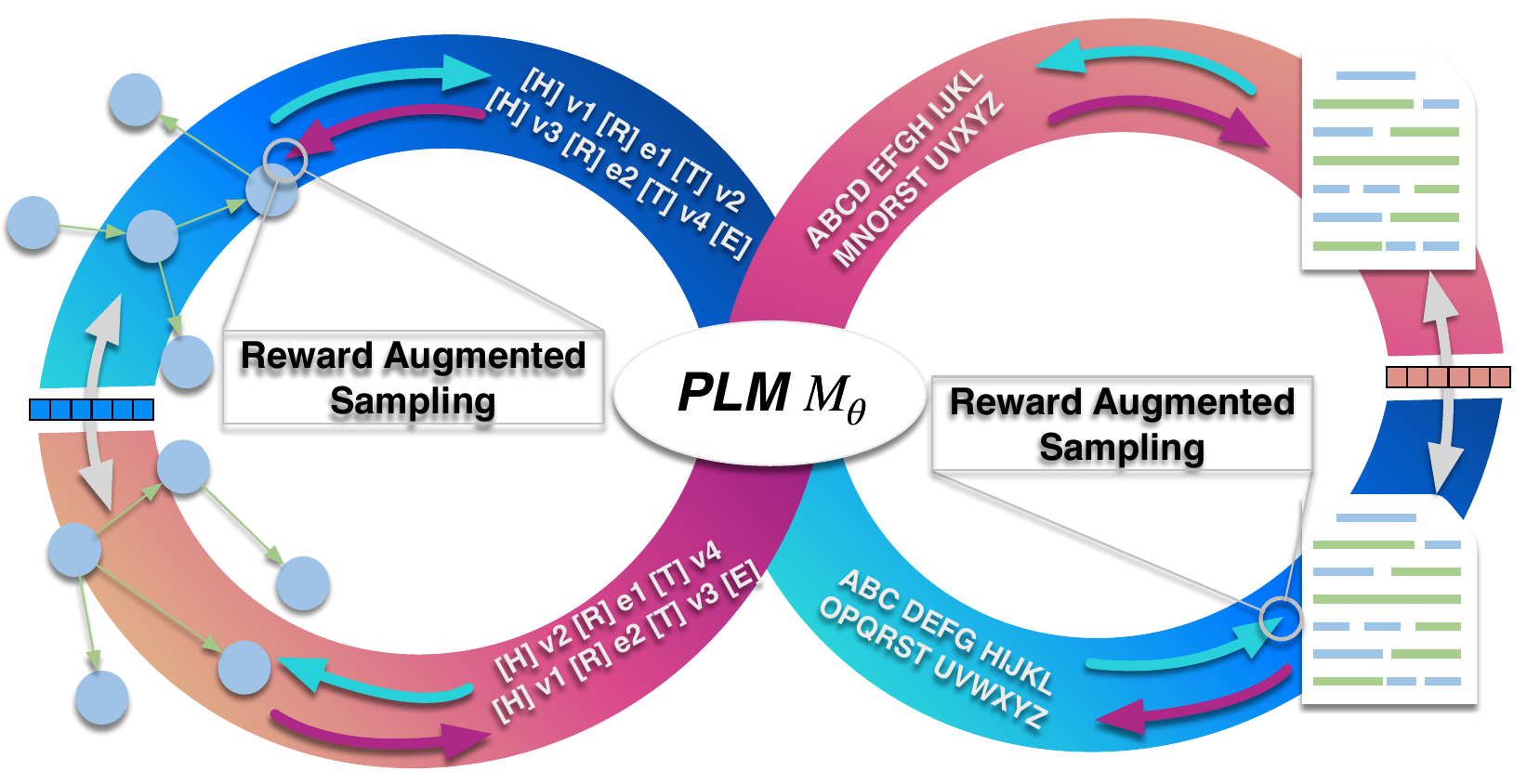}
    \caption{The overall architecture of \textit{INFINITY}. The cycle of blue arrows illustrates the direction of G2T task while the cycle of magenta arrows illustrates the direction of T2G task.}
    \label{fig:infinity}
\end{figure}

The overall architecture of \textit{INFINITY} is shown in Figure~\ref{fig:infinity}, which is shaped like $\infty$. The framework iteratively back-translates between graph dataset and text corpus, where the vocabulary embeddings and the seq2seq-based PLM denoted by $M_{\theta}$ are the same for G2T and T2G tasks. For simplicity and with a slight abuse of notation, we use the same symbol $M_{\theta}(\cdot)$ to represent the sequence generating function of the PLM, whether its output is discrete or continuous. The training process of \textit{INFINITY} consists of two parts.

\textbf{Graph $\rightarrow$ Text $\rightarrow$ Graph.} In this part (the cycle of blue arrows), we first linearize the original graph into a triple sequence with special tokens. The linearized graph $g$ is then fed to the encoder of $M_{\theta}$ for further training. Afterward, the decoder of $M_{\theta}$ generates a synthetic text sequence denoted by $M_{\theta}(g)$ as the intermediate result. It is worth noting that we use the text embedding generated from $M_{\theta}$ instead of the discrete tokens as the input for the PLM's encoder in the following step. Finally, the PLM receives the synthetic text embedding and generates a back-translated graph, which is used to align the original one through maximum likelihood estimation. Ideally, the back-translated graph should mimic the original graph $g$.


\textbf{Text $\rightarrow$ Graph $\rightarrow$ Text.} Similarly, for the other direction (the cycle of magenta arrows), the PLM $M_{\theta}$ first generates a synthetic graph sequence denoted by $M_{\theta}(t)$ from a text $t$. Then, we also use the embedding of the synthetic graph sequence and feed it to $M_{\theta}$. In the end, $M_{\theta}$ generates a back-translated text from the synthetic graph embedding for the original text $t$ to train the parameters. We expect the back-translated text to be roughly the same as the original text.

Different from the unsupervised back-translation~\citep{lample2018phrase,guo2020cyclegt}, \textit{INFINITY} only employs one neural network, i.e. the PLM $M_{\theta}$ for these two tasks. In this way, the parameters of the framework are greatly reduced. The only PLM can observe the original graphs and texts at the same time, which is easier for information sharing and model training. In summary, G2T and T2G can be optimized simultaneously in the proposed \textit{INFINITY} with synthetic parallel pairs $(t, M_{\theta}(t))$ and $(g, M_{\theta}(g))$. The objective is as follows:


\begin{equation}
    \begin{aligned}
        \mathcal{L} = \; & \mathbb{E}_{g \in \mathcal{G}}[-\log P(g|M_{\theta}(g))] + \\ 
        & \mathbb{E}_{t \in \mathcal{T}}[-\log P(t|M_{\theta}(t))]
    \end{aligned}
\end{equation}

\subsection{Reward Augmented Training Loss}
As a framework to solve the problem of bidirectional sequence generation, we need to consider how to retain more structural information in graphs as much as possible without introducing additional parameters. In detail, graph linearization strategy hinders seq2seq-based PLM from capturing graph structure with maximum likelihood estimation (MLE) since MLE suffers from the exposure bias problem~\citep{bengio2015scheduled}. To this end, we adopt reward augmented maximum likelihood (RML) ~\citep{norouzi2016reward} which combines the primary form of MLE and the maximum reward expectation in reinforcement learning (RL). In this way, our training process can make rewards one of the training targets under the framework of MLE, which considers the structure of graphs and the order of texts. According to RML, the \textit{exponentiated payoff distribution} connects MLE and RL objectives, and it can be easily incorporated into MLE-based training. In our framework, we define a distribution in the augmented space for graph dataset $\mathcal{G}$ as follows:

\begin{align}\label{graph_distribution}
    q(\tilde{g}|g;\tau) = \frac{exp(r(\tilde{g},g)/\tau)}{\sum_{\tilde{g} \in \tilde{\mathcal{G}}}exp(r(\tilde{g},g)/\tau)},
\end{align}
where $\tilde{g} \in \tilde{\mathcal{G}}$ is the output hypothesis (possible generated sequence) of $g$, $r(\tilde{g},g)$ denotes the reward function such as BLEU or F1 score, $\tau$ is the temperature to control the degree of regularization, and $\tau > 0$. Now, we modify the MLE-based objective function to:

\begin{align}\label{graph_raml}
    \mathcal{L}_{RML}^{G} = \mathbb{E}_{g \in \mathcal{G}}[-\sum_{\tilde{g} \in \tilde{\mathcal{G}}}q(\tilde{g}|g;\tau)\log p(\tilde{g}|M_{\theta}(g))].
\end{align}

In $\mathcal{L}_{RML}^{G}$, the predictive probability of the outputs in the original loss can be smoothed using their corresponding rewards with the distribution $q(\tilde{g}|g;\tau)$. For symmetry, RML can also be extended to our text corpus $\mathcal{T}$ similarly, which is as follows:

\begin{align}\label{text_distribution}
    q(\tilde{t}|t;\tau) = \frac{exp(r(\tilde{t},t)/\tau)}{\sum_{\tilde{t} \in \tilde{\mathcal{T}}}exp(r(\tilde{t},t)/\tau)},
\end{align}

\begin{align}\label{text_raml}
    \mathcal{L}_{RML}^{T} = \mathbb{E}_{t \in \mathcal{T}}[-\sum_{\tilde{t} \in \tilde{\mathcal{T}}}q(\tilde{t}|t;\tau)\log p(\tilde{t}|M_{\theta}(t))],
\end{align}
where $\tilde{t} \in \tilde{\mathcal{T}}$ is the output hypothesis of $t$. However, experiments show that the strategy will not significantly improve the performance when applied to text sequences. In addition, existing unsupervised models such as ~\citet{guo2020cyclegt} \textbf{cannot employ RML} for graph extraction, which is defined as a relational classification problem rather than a sequence generation problem. 

The system of RML is simple and computationally efficient. One only needs to sample possible outputs $\tilde{\mathcal{G}}$ and $\tilde{\mathcal{T}}$ from their corresponding exponentiated payoff distribution before training. According to~\citet{norouzi2016reward}, it is difficult to sample with BLUE or F1 score since the distribution is intractable to compute. Thus, we adopt the importance sampling method with the distribution of hamming distance between the original sequence and its hypothesis. More theoretical details can be found in \citet{norouzi2016reward}.



\subsection{Training and Inference Details}

In \textit{INFINITY}, G2T and T2G tasks are jointly trained thanks to the shared parameters. Compared with unsupervised machine translation, our model does not train a language model with the denoising auto-encoder objective on the two tasks due to the shared vocabulary of PLM. As a result, we optimize the loss function:

\begin{align}
    \mathcal{L} = \mathcal{L}_{RML}^{G} + \mathcal{L}_{RML}^{T}.
\end{align}

The detailed training process of unsupervised \textit{INFINITY} is provided in Algorithm~\ref{alg:training}. In our implementation, we use T5-base~\citep{raffel2019exploring} as the PLM since T5 is based on transformer and can handle multiple tasks well. We prepend graph prefix \underline{\textit{Graph: }} to the linearized graph sequence for G2T task and text prefix \underline{\textit{Text: }} for T2G task. In order to speed up the convergence of training, when generating synthetic intermediate outputs of texts, we discard embeddings of illegal tokens including $[H], [R], [T]$, and $[E]$ for the G2T task, which will not be fed to the encoder of the PLM in the following step. During the inference stage, we leverage the beam search to generate texts and linearized graphs. Additionally, for the T2G direction, we adopt the same heuristic rules recommended in prior work~\citep{ye2021contrastive} to generate reasonable linearized graphs, where the special token $[R]$ (relation) should be followed by $[H]$ (head entity).

\begin{algorithm}[htb]
\caption{Training Unsupervised \textit{INFINITY}}
\label{alg:training}
\begin{algorithmic}[1] 
\STATE Initiate parameters of PLM $M_\theta^{(1)}$;
\STATE Obtain the distribution of $q(\tilde{g}|g;\tau)$ from $\mathcal{G}$ according to Equation~\ref{graph_distribution};
\STATE Obtain the distribution of $q(\tilde{t}|t;\tau)$ from $\mathcal{T}$ according to Equation~\ref{text_distribution};
\FOR{i = 1 to N}
\STATE $\tilde{\mathcal{G}}^{(i)} \leftarrow M_\theta^{(i)}(M_\theta^{(i)}(\mathcal{G}))$;
\STATE $\tilde{\mathcal{T}}^{(i)} \leftarrow M_\theta^{(i)}(M_\theta^{(i)}(\mathcal{T}))$;
\STATE Compute $\mathcal{L}_{RML}^{G}$ using $\tilde{\mathcal{G}}^{(i)}$ and $\mathcal{G}$ according to Equation~\ref{graph_raml};
\STATE Compute $\mathcal{L}_{RML}^{T}$ using $\tilde{\mathcal{T}}^{(i)}$ and $\mathcal{T}$ according to Equation~\ref{text_raml};
\STATE $\mathcal{L} \leftarrow \mathcal{L}_{RML}^{G} + \mathcal{L}_{RML}^{T}$;
\STATE Fine-tune $M_\theta^{(i)}$ with $\mathcal{L}$ and obtain $M_\theta^{(i+1)}$;
\ENDFOR
\STATE \textbf{return} $M_\theta^{(N+1)}$;
\end{algorithmic}
\end{algorithm}

\section{Experiments}
This section conducts a series of experiments to evaluate the performance of \textit{INFINITY}. We first introduce the datasets and baselines, then we provide the comparison results. At last, we implement extensive analytical experiments, including ablation analysis and case study.

\subsection{Datasets}
Since our task is unsupervised, datasets with external information except for graphs and texts are not in our consideration. Thus, we select WebNLG (2017)~\cite{gardent2017webnlg} and GenWiki~\cite{jin2020genwiki} as our benchmarks, which can evaluate G2T and T2G models at the same time. WebNLG is widely used in text generation and relation extraction, where each graph contains about 2 to 7 triples. GenWiki is a new resource for unsupervised G2T generation, and we select two large domains (i.e., Sports and Games) of GenWiki. Tabel~\ref{statistics} presents the detailed statistics of these two datasets.

\begin{table}[!h]
\centering
\small
\begin{tabular}{c|cccc}
\hline
\textbf{Dataset} & Train  & Valid & Test   & Relation Types \\ \hline
\textbf{WebNLG}  & 18,102 & 872   & 1,862  & 373            \\
\textbf{GenWiki} & 48,020 & 1,000 & 10,000 & 250            \\ \hline
\end{tabular}
\caption{Statistics of benchmarks.}
\label{statistics}
\end{table}

\subsection{Baselines}
\subsubsection{Supervised Baselines}
The intended application of \textit{INFINITY} is in unsupervised scenarios. Thus, only related methods are considered. For G2T, we compare our model with a wide selection of PLM-free and PLM-based methods. PLM-free models include \emph{StrongNeural}, \emph{BestPlan}~\citep{moryossef2019step}, \emph{GraphWriter}~\citep{koncel2019text}, and \emph{Planner}~\citep{zhao2020bridging}, where \emph{BestPlan} and \emph{Planner} design different planners to order triples before linearization. PLM-based models include~\emph{T5-base} and \emph{T5-large}~\citep{ribeiro2020investigating}. As to T2G, we choose \emph{OnePass}~\citep{wang2019extracting} and a state-of-the-art triple extraction model \emph{CGT}~\citep{ye2021contrastive} as our baselines. Moreover, we also implement a supervised 
version of \textit{INFINITY} with aligned graph-text pairs, which serves as a reference for the upper bound of our unsupervised model. The supervised loss is:

\begin{align}
    \mathcal{L}^{sup}=\mathbb{E}_{(g,t) \in \mathcal{G} \times \mathcal{T}}[-\log P(g|t)-\log P(t|g)].
\end{align}

\subsubsection{Unsupervised Baselines}
Due to the limited research on unsupervised joint training, we selected all unsupervised models as baselines. \emph{Rule-Based}~\citep{schmitt2020unsupervised} employs a heuristic algorithm to extract facts and concatenate text of each triplet. \emph{Graph-Text Back Translator (GT-BT)}~\citep{schmitt2020unsupervised} adopts a series of denoising methods and applies a back-translation model with a POS tagger as external tool. \emph{CycleGT}~\citep{guo2020cyclegt} jointly trains both tasks via cycle training, where the T2G is simplified to the relation classification task with given entities.

\subsection{Training Settings and Evaluation Metrics}
We employ Adam as the optimizer. The beam size is set to 4 for both tasks. The learning rate is set to 1e-4. For G2T, we adopt several widely used automatic metrics, i.e., BLEU~\citep{papineni2002bleu}, Meteor~\citep{banerjee2005meteor}, and CIDEr~\citep{vedantam2015cider}. BLEU and Meteor consider precision, recall, or F-score between generated and ground truth texts while CIDEr calculates the TF-IDF weights for each $n$-gram. For T2G, we use the micro F1 score to evaluate the quality of the generated triples. F1 results of entities and triples are provided. We select part of the above metrics to show due to space limitations.


\subsection{WebNLG Results}
\subsubsection{G2T Results} 
Table~\ref{cmp_webnl_g2t} presents the results of G2T task on the WebNLG dataset. For fairness, we report the results of \textit{INFINITY} \textbf{without applying RML to texts}, which is also analyzed in the ablation section. It can be seen that our proposed method outperforms all other unsupervised baselines. The BLEU score of \textit{INFINITY} is 2 points higher than CycleGT, which is even better than the level of some supervised models. Moreover, the performance of the supervised \textit{INFINITY} is on par with T5-base and T5-large, and the supervised version can even deal with the T2G problem, which can be attributed to the power of PLM and the joint optimization for the shared latent space.

\begin{table}[!ht]
\centering
\small
\setlength{\tabcolsep}{2.0mm}{\begin{tabular}{lccc}
\hline
                    & \textbf{BLEU} & \textbf{METEOR} & \textbf{CIDEr} \\ \hline
\multicolumn{4}{l}{\textbf{Supervised Models (G2T)}}                   \\
StrongNeural        & 46.5          & 0.39           & 2.87           \\
BestPlan            & 47.4          & 0.39           & 2.69           \\
GraphWriter         & 45.8          & 0.36           & 3.14           \\
Planner             & 52.9          & 0.45           & 3.72           \\
T5-base             & 59.1          & 0.44           & 4.02           \\
T5-large            & 59.3          & 0.44           & 4.03  \\
Supervised \textit{INFINITY} & 58.8 & 0.44           & 3.99           \\
\multicolumn{4}{l}{\textbf{Unsupervised Models (Given Entities / External Tools)}}      \\
Rule-Based          & 18.3          & 0.34           & -              \\
GT-BT               & 37.7          & 0.36           & -              \\
CycleGT             & 55.5          & 0.44           & 3.81           \\
\multicolumn{4}{l}{\textbf{Unsupervised Models}}                       \\
\textit{INFINITY}   & 58.0          & 0.44           & 3.89           \\ \hline
\end{tabular}}
\caption{G2T performance comparisons of different models on WebNLG dataset. CIDEr results and corresponding codes are not provided in Rule-Based and GT-BT.}
\label{cmp_webnl_g2t}
\end{table}

\subsubsection{T2G Results} 
For the T2G task, it should be mentioned that the compared three unsupervised models RuleBased, GT-BT, and CycleGT, are given entities as a relation classification task, so they have a 100\% F1 score of entities naturally and cannot employ RML loss for graph sequences. As can be seen from Table~\ref{cmp_webnl_t2g}, our model's F1 (triple) score is $61.7$, which is superior to all other unsupervised models under the circumstance that all entities are unknown. Rule-Based model cannot extract any triples. Our supervised \textit{INFINITY} shows better results than the unsupervised one in terms of entity recognition, whereas its performance is inferior to other supervised methods since our model only uses the T5-base PLM and does not design other sophisticated modules.

\begin{table}[htb]
\centering
\small
\setlength{\tabcolsep}{2.0mm}{\begin{tabular}{lcc}
\hline
                    & \textbf{F1 (entity)} & \textbf{F1 (triple)} \\ \hline
\multicolumn{3}{l}{\textbf{Supervised Models (T2G)}}              \\
OnePass             & NA                   & 66.2                 \\
CGT                 & NA                   & 83.4                 \\
Supervised \textit{INFINITY} & 95.0        & 59.3                 \\
\multicolumn{3}{l}{\textbf{Unsupervised Models (Given Entities / External Tools)}} \\
Rule-Based          & 100.0                & 0.0                  \\
GT-BT               & 100.0                & 39.1                 \\
CycleGT             & 100.0                & 58.4                 \\
\multicolumn{3}{l}{\textbf{Unsupervised Models}}                  \\
\textit{INFINITY}   & 93.9                 & 61.7                 \\ \hline
\end{tabular}}
\caption{T2G performance comparisons of different models on WebNLG dataset. NA means the model is not applicable to extract entities.}
\label{cmp_webnl_t2g}
\end{table}

\subsection{GenWiki Results}
Unlike the WebNLG dataset, GenWiki is specially collected for unsupervised G2T tasks, where graph elements do not necessarily exist in the text. Moreover, the entities extracted from the text are also not necessarily contained in the ground truth graph, which makes it challenging to generate informative outputs. Hence, some supervised baselines are not applicable to this dataset. Since the codes of Rule-Based and GT-BT~\citep{schmitt2020unsupervised} are not provided, we use our implemented Rule-Based model as the baseline. In Table~\ref{cmp_genwiki}, our proposed method shows better results than GraphWriter and Rule-Based model, but the BLEU value of \textit{INFINITY} is lower than CycleGT. The reason is that CycleGT has known all tokens of entities and relation types for T2G task, which can be used as external information to achieve better performance during the training process. As a result, \textit{INFINITY} can only generate the tokens of entities and relations that appear in the original texts. In other words, our model may substitute the ground truth tokens with other words but remain the similar meanings. For example, the original relation \textit{birthYear} may be predicted as \textit{birthDay} in \textit{INFINITY}.

\begin{table}[htb]
\centering
\small
\setlength{\tabcolsep}{0.4mm}{\begin{tabular}{lccc}
\hline
\multirow{2}{*}{}   & \multicolumn{2}{l}{\textbf{G2T}} & \textbf{T2G}         \\
                    & \textbf{BLEU}  & \textbf{CIDEr}  & \textbf{F1 (triple)} \\ \hline
\multicolumn{4}{l}{\textbf{Supervised Models}}                                \\
GraphWriter         & 29.7          & 2.68            & NA                    \\
T5-base             & 45.7          & 3.74            & NA                    \\
T5-large            & 47.1          & 3.74            & NA                    \\
Supervised \textit{INFINITY} & 43.6 & 3.44            & 33.8                  \\
\multicolumn{4}{l}{\textbf{Unsupervised Models (Given Entities / External Tools)}}             \\
Rule-Based (our implementation)& 13.9 & 1.26          & 0.0                   \\
CycleGT             & 38.5          & 3.50            & 34.2                \\
\multicolumn{4}{l}{\textbf{Unsupervised Models}}                              \\
\textit{INFINITY}   & 34.3          & 2.50            & 23.4                \\ \hline
\end{tabular}}
\caption{G2T and T2G performance comparisons of different models on GenWiki dataset.}
\label{cmp_genwiki}
\end{table}

\subsection{Detailed Analysis}
\subsubsection{Ablation Study} We use the WebNLG dataset for ablation analysis. As shown in Table~\ref{ablation_webnlg}, the supervised \textit{INFINITY} shows the best results on the G2T task while the performance of \textit{INFINITY} without reward augmented losses (w/o RML) is worse than any other versions, especially for T2G. Applying reward augmented loss to both text and graph makes the model capture more order and structural information in the datasets, and it obtains significant improvement. We also evaluate variants that only adopt one side reward augmented loss. \textit{INFINITY} with RML for graph demonstrates the best performance except for the supervised one. This is because the PLM itself performs well on texts, and the improvement of RML for text is limited. Therefore, we use the version with RML for graph as our final reported model.

\begin{table}[htb]
\centering
\small
\setlength{\tabcolsep}{1.6mm}{\begin{tabular}{lccc}
\hline
                                   & \multicolumn{2}{l}{\textbf{G2T}} & \multicolumn{1}{l}{\textbf{T2G}} \\
                                   & \textbf{BLEU}  & \textbf{CIDEr}  & \textbf{F1 (triple)}    \\ \hline
Supervised \textit{INFINITY}       & \textbf{58.8}  & \textbf{3.99}   & 59.3           \\
w/o RML                            & 54.3  & 3.58   & 51.5           \\
w. RML for text \& graph           & 57.3  & 3.89   & 59.7           \\
w. RML for text                    & 56.2  & 3.67   & 53.8           \\
w. RML for graph (ours)            & 58.0  & 3.89   & \textbf{61.7}           \\\hline
\end{tabular}}
\caption{Ablation analysis on WebNLG dataset. The version with RML for graph is used as our reported results.}
\label{ablation_webnlg}
\end{table}

\begin{table*}[!ht]
\centering
\small
\begin{tabular}{l|cccc|cccc}
\hline
                          & \multicolumn{4}{c|}{\textbf{WebNLG}}                                 & \multicolumn{4}{c}{\textbf{GenWiki}}                                \\ \cline{2-9} 
                          & \multicolumn{2}{c}{\textbf{G2T}} & \multicolumn{2}{c|}{\textbf{T2G}} & \multicolumn{2}{c}{\textbf{G2T}} & \multicolumn{2}{c}{\textbf{T2G}} \\  
                          & BLEU            & CIDEr          & F1 (entity)     & F1 (triple)     & BLEU            & CIDEr          & F1 (entity)     & F1 (triple)    \\ \hline
WebNLG.G $\times$ GenWiki.T & 34.8           & 2.04          & 89.1            & 45.2            & 21.6           & 1.41          & 59.2            & 1.2            \\
WebNLG.T $\times$ GenWiki.G & 45.6           & 2.82          & 91.9            & 19.5            & 16.1           & 1.13          & 65.6            & 9.1            \\ \hline
WebNLG                      & 58.0           & 3.89          & 93.9            & 61.7            & NA             & NA            & NA              & NA             \\
GenWiki                     & NA             & NA            & NA              & NA              & 34.3           & 2.50          & 97.0            & 23.4           \\
\hline
\end{tabular}
\caption{Analysis of cross learning on WebNLG and GenWiki datasets. $dataset.G$ means the graph data in $dataset$ and $dataset.T$ denotes the text corpus in $dataset$. The last two rows are the results of training with the graphs and texts on a single dataset, where they are only evaluated on their corresponding benchmark.}
\label{cross_learning}
\end{table*}

\begin{table*}[!ht]
\centering
\small
\renewcommand\arraystretch{1.2}
\begin{tabular}{l|l}
\hline
                         & \textbf{Instance} \\ \hline
\textbf{G.T. Text}       & \textit{Arlington in Texas is located at 184.0 metres above sea level and has a total area of 258.2 square kilometres.} \\ \cline{2-2} 
\textbf{Gen. Text}  & Arlington, Texas is 184.0 above sea level and has a total area of 258.2 square kilometres. \\ \cline{2-2} 
\textbf{G.T. Graph}      & \begin{tabular}[c]{@{}l@{}}$[H]$ Arlington Texas $[R]$ elevation Above The Sea Level $[T]$ 184.0 \\ $[H]$ Arlington Texas $[R]$ area Total $[T]$ 258.2 square kilometres $[E]$\end{tabular} \\ \cline{2-2} 
\textbf{Gen. Graph} & \begin{tabular}[c]{@{}l@{}}$[H]$ Arlington Texas $[R]$ elevation Above The Sea Level $[T]$ 184.0 \\ $[H]$ Arlington Texas $[R]$ area Total $[T]$ 258.2 square kilometres $[E]$\end{tabular} \\ \hline
\textbf{G.T. Text}       & \begin{tabular}[c]{@{}l@{}}\textit{The Aston Martin V8, manufactured by Aston Martin, is related to the Aston Martin DBS and was succeeded} \\ \textit{by the Aston Martin Vantage. Its engine volume is 5.3 litres. and it is assembled at Newport Pagnell.}\end{tabular} \\ \cline{2-2} 
\textbf{Gen. Text}  & \begin{tabular}[c]{@{}l@{}}The Aston Martin V8, with a 5.3 litre engine, is a related transport vehicle to the Aston Martin DBS. \\ It is the successor to the Newport Pagnell Aston Martin Vantage.\end{tabular} \\ \cline{2-2} 
\textbf{G.T. Graph}      & \begin{tabular}[c]{@{}l@{}}$[H]$ Aston Martin V8 $[R]$ related Mean Of Transportation $[T]$ Aston Martin DBS \\ $[H]$ Aston Martin DBS $[R]$ successor $[T]$ Aston Martin Vantage \\ $[H]$ Aston Martin V8 $[R]$ assembly $[T]$ Newport Pagnell \\ $[H]$ Aston Martin V8 $[R]$ engine $[T]$ 5.3 litres \\ $[H]$ Aston Martin V8 $[R]$ manufacturer $[T]$ Aston Martin $[E]$\end{tabular} \\ \cline{2-2} 
\textbf{Gen. Graph} & \begin{tabular}[c]{@{}l@{}}$[H]$ Aston Martin V8 $[R]$ manufacturer $[T]$ Aston Martin \\ $[H]$ Aston Martin DBS $[R]$ succeeded By $[T]$ Aston Martin Vantage \\ $[H]$ AstonMartin V8 $[R]$ engine Volume $[T]$ 5.3 litres \\ $[H]$ Aston Martin assembly location $[R]$ Newport Pagnell $[E]$\end{tabular} \\ \hline
\end{tabular}
\caption{Case Study on WebNLG dataset. G.T. means \textit{ground truth} and Gen. means \textit{generated}.}
\label{case_webnlg}
\vspace{-10pt}
\end{table*}

\subsubsection{Analysis of Cross Learning} 
As mentioned earlier, we assume $\mathcal{T}$ and $\mathcal{G}$ share the same latent content. In the same dataset, $\mathcal{T}$ and $\mathcal{G}$ have the same domain knowledge, whereas different datasets can only share the language. In the latter case, to analyze the scalability of \textit{INFINITY}, we introduce cross learning where we only use the graph (or text) data of WebNLG and text (or graph) corpus of GenWiki for training. Table~\ref{cross_learning} shows the results, where $dataset.G$ means the graph in $dataset$ while $dataset.T$ denotes the text in $dataset$. We can see \textit{INFINITY} works well under the setting of cross learning, which cannot be accomplished by other unsupervised models such as CycleGT since they require entities and relation types for both tasks. However, the T2G performance of GenWiki is worse than WebNLG because the tokens of relations and texts rarely overlap in GenWiki. In summary, \textit{INFINITY} provides a low-resource approach to deploy on different datasets for application. For example, in the absence of a corresponding graph corpus, we can use public knowledge graphs to train \textit{INFINITY} model so as to extract graph triples from any given English literature.

\subsubsection{Case Study and Error Analysis} To analyze the generation performance and drawbacks of \textit{INFINITY}, we select two representative instances shown in Table~\ref{case_webnlg}, where the ground truth and generated sequences are provided. As to the first case, the generated text is consistent with the ground truth, with only slight differences, and the generated triples are exactly the same as the real ones. The second instance contains two sentences and five triples. The order of the generated text is inconsistent with the original text, and there are some semantic errors. The generated triples are all reasonable but miss the first fact. The boundary of the last generated triple is wrong, where $[T]$ is missing.

\section{Conclusion and Future Work}
In this paper, we propose \textit{INFINITY}, a simple unsupervised approach to graph-text mutual conversion. The key idea of \textit{INFINITY} is to utilize one seq2seq-based PLM to converse graphs and texts from each other with the framework of back-translation. Unlike existing unsupervised methods, our model requires no additional external information or tools beyond the non-parallel graph and text corpus, so it is easy to be quickly deployed to industrial scenarios. Experimental results show that \textit{INFINITY} achieves promising results compared to state-of-the-art baselines. For future work, we plan to explore the capability of prompt learning by appealing to precise controls over different layers in PLMs.

\clearpage

\bibliography{anthology,custom}
\bibliographystyle{acl_natbib}

\end{document}